\documentclass[10pt]{article}
\usepackage{aaai}
\usepackage{times}
\usepackage{helvet}
\usepackage{courier}
\usepackage[english]{babel}
\usepackage[utf8x]{inputenc}
\usepackage{amsmath}
\usepackage[pdftex]{graphicx}
\usepackage[hyphens]{url}
\usepackage{subcaption}
\begin{document}

\title{Bio-YODIE: A Named Entity Linking System for Biomedical Text}

\author{Genevieve Gorrell\\
{University of Sheffield, UK}\\
\texttt{g.gorrell@sheffield.ac.uk}
\And
Xingyi Song\\
{University of Sheffield, UK}\\
\texttt{x.song@sheffield.ac.uk},
\And
Angus Roberts,\\
{King's College London, UK}\\
\texttt{angus.roberts@kcl.ac.uk}}

\nocopyright
\maketitle
\begin{abstract}
\begin{quote}
Ever-expanding volumes of biomedical text require automated semantic annotation techniques to curate and put to best use. An established field of research seeks to link mentions in text to knowledge bases such as those included in the UMLS (Unified Medical Language System), in order to enable a more sophisticated understanding. This work has yielded good results for tasks such as curating literature, but increasingly, annotation systems are more broadly applied. Medical vocabularies are expanding in size, and with them the extent of term ambiguity. Document collections are increasing in size and complexity, creating a greater need for speed and robustness. Furthermore, as the technologies are turned to new tasks, requirements change; for example greater coverage of expressions may be required in order to annotate patient records, and greater accuracy may be needed for applications that affect patients. This places new demands on the approaches currently in use. In this work, we present a new system, Bio-YODIE, and compare it to two other popular systems in order to give guidance about suitable approaches in different scenarios and how systems might be designed to accommodate future needs.
  \end{quote}
\end{abstract}

\section{Introduction}


The recent explosion in availability of textual materials has created opportunities to benefit from large-scale automated text analysis, and in no domain is this more true than the biomedical domain, where volumes of research literature are increasing exponentially~\cite{fujiwara2015}, patient records are surging into electronic form, and the potential for real world impact is extensive. Semantic annotation aims to facilitate work such as epidemiology, clinician decision support and research case sample identification through the ability to dereference mentions in text to concepts in a knowledge base. This allows more sophisticated inference to be performed than through text search alone, such as identifying mentions of all antipsychotic drugs.

Named entity linking (NEL) is an established field of research that aims to achieve this goal. It is at an earlier stage of development in the biomedical domain than in the general domain, perhaps in part because of the success of MetaMap~\cite{aronson2006metamap} in providing a result suitable for many needs. MetaMap was designed to provide first draft topic labels for PubMed articles, for manual correction, which is a different task to accurately dereferencing mentions in text for applications potentially affecting clinical decisions. This repurposing warrants appropriate evaluation. Additionally, as larger volumes of text require processing, computational speed becomes increasingly important, so this too must be evaluated.

Named entity recognition (NER) describes the task of finding parts of text that appear to be the name of something, and is a valuable contribution in its own right. Named entity linking can be seen as a second step to this, in which the names, once found, are looked up in the database and a match located, and this is the approach taken in many systems, including MetaMap. It might be seen as a "pull" approach, in that linking is driven by seeking a match for the mention already found in the text. An alternative approach is to begin with a list of all possible names of entities of interest (a "gazetteer"), and look for matching strings in the text. It might be said in that case that the links are "pushed" onto the document by the gazetteer. MetaMapLite~\cite{demner2017metamap} is an example of such a system. In the general domain, we have seen that approaches based on an initial named entity recognition (NER) stage have different properties when compared with approaches based on finding string matches in a gazetteer, and different strengths and weaknesses. Where text is ill-formed, perhaps with poor capitalization, such as might be found in rapidly penned patient notes, NER may perform poorly, becoming a bottle-neck for the performance of the system as a whole. Gazetteer-based approaches have the potential to be faster and more robust, though they lack the ability to identify "unknowns"; that is, entities that aren't currently in the database but perhaps ought to be. However, for many applications this isn't needed, and where it is, a gazetteer-based approach may be supplemented with an NER approach, retaining the advantages of both rather than making one a dependency of the other.

In both cases, it is important to introduce some flexibility into term matching, as there are likely to be synonyms and common spelling errors that aren't included on the official list of names for an entity, such as that contained in the UMLS (Unified Medical Language System, a popular compendium of medical vocabularies). In the "pull" approach, having found a mention in text, matching candidates are sought in the database, and some flexibility may be employed in how this is done. The string found in text may be permuted, for example by trying different morphological word endings, before searching in the database. In the "push" case, the labels in the database are permuted, thus lengthening the gazetteer, allowing it to match more terms. An advantage to permuting the labels in the database is that it can be done off-line, ahead of time, saving time and potentially increasing robustness at runtime, albeit at the cost of an increased memory requirement for the pre-calculated data resource. In this paper we discuss the practical impact of coverage extension strategies on the systems discussed.

The second way in which performance can be improved is through the quality of disambiguation. Where a string in text might refer to multiple entities (for example "OD" might refer to "overdose", "once daily" or "osteochondritis dissecans"), one interpretation must be chosen. A limited domain containing relatively unique terms faces less of an issue with homonyms. Disambiguation work in the biomedical domain has been mainly confined to acronyms and abbreviations. However, as the coverage and richness of vocabularies increases, we can expect that the need for disambiguation will increase, and that we may have more need to learn from work in the general domain. Certainly there is a need for evaluation, to scope the problem and assess the success of strategies.

In this work we present a new biomedical named entity linking system arising within the GATE natural language processing ecosystem, called Bio-YODIE. We contrast Bio-YODIE with MetaMap and MetaMapLite in order to elucidate the strengths and weaknesses of each approach, and draw generalizations about which approach might be best suited in a variety of situations. We demonstrate that Bio-YODIE performs competitively. In the next section we describe Bio-YODIE, before moving on to evaluation.

\section{The Bio-YODIE Biomedical Named Entity Linking System}

Bio-YODIE has emerged in parallel evolution with MetaMapLite, implementing broadly the same philosophy. Based on GATE~\cite{cunningham2011text} technology, YODIE~\cite{gorrell2015using} was originally a general domain system, dereferencing mentions to DBpedia~\cite{auer2007dbpedia}. Bio-YODIE is the biomedical version of the system, though inheriting a research history from the general domain. It differs from MetaMapLite in that disambiguation has always been a priority in the YODIE project, and remains so in Bio-YODIE. Bio-YODIE has been integrated into CogStack~\cite{jackson2018cogstack} and is in active use in large scale clinical contexts.

Bio-YODIE comprises two main components. Firstly, the resource preparation step processes the UMLS and other informational resources required at runtime into an efficient form in which as much work has been done as possible in advance to minimize processing at runtime. Secondly, the pipeline itself applies annotations to the documents that include a UMLS Concept Unique Identifier (CUI) along with other pertinent information from the UMLS. Resource preparation takes the form of scripts that can be distributed in order that others may prepare resources from their own licensed UMLS download on the subset of their choice. The resources comprise a gazetteer and a database allowing entities to be retrieved by label (valid name mention in text). A number of scores and other relevant information are made available in the entry for each entity. The scripts can be used to generate resources in any language for which there is UMLS content. They also include a type list, so that only the required semantic types (for example "Sign or Symptom", "Finding") need be included.

\begin{figure}[h]
\centering
\includegraphics[width=0.5\textwidth]{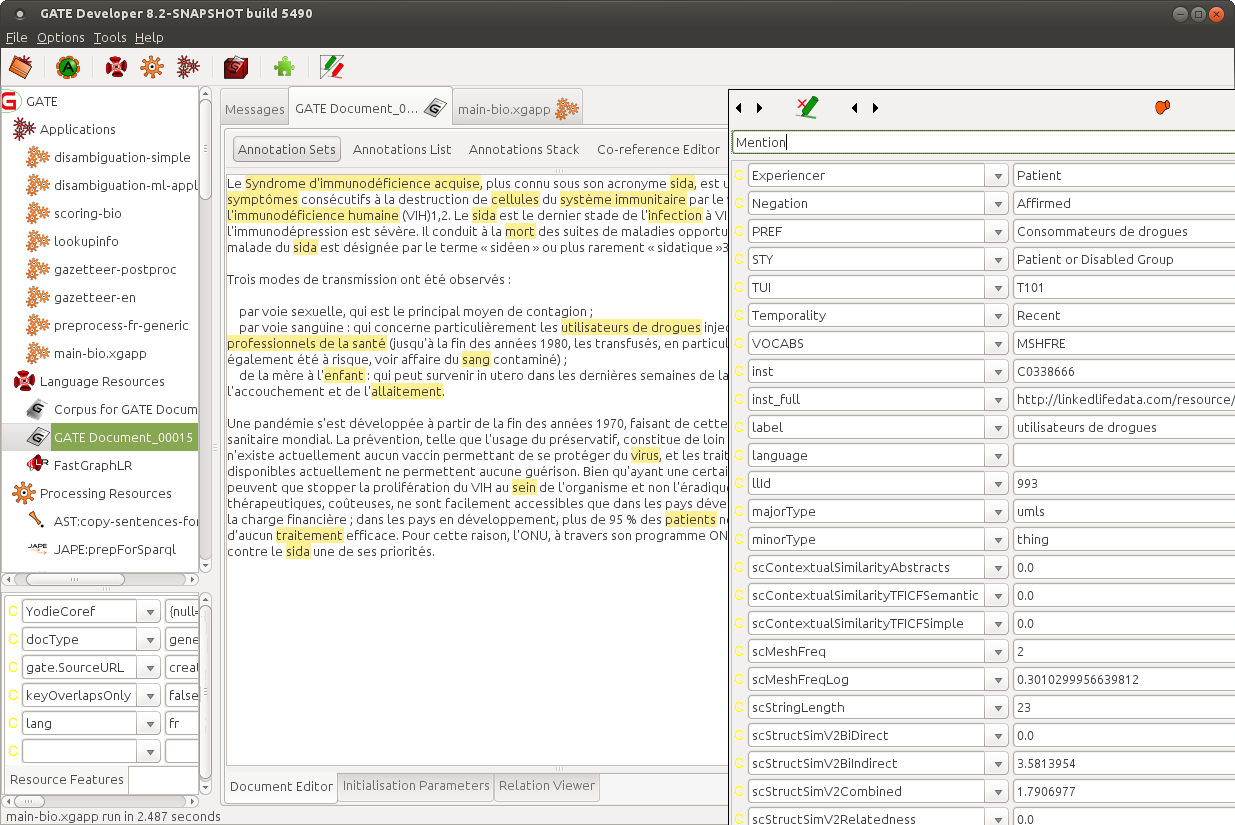}
\caption{Result of running Bio-YODIE on example French medical text}
\label{fig:bioyodie}
\end{figure}

The pipeline begins by running a gazetteer to locate the terms that may indicate an entity mention. After using a stop list to remove low precision terms, these mentions are then used to retrieve all the possible candidates for that term. A number of scores are then used to pick the most likely; this is discussed in more detail in the disambiguation section below. The system can be run using the GATE Developer GUI, or from the command line; for example GCP (GATE Cloud Processor) is routinely used to run Bio-YODIE in multiple threads over the CRIS database of 30 million patient notes. Figure~\ref{fig:bioyodie} shows the result of Bio-YODIE having been run over an example document in French. The pane to the right is a pop-up pane displaying information about the referent assigned to "utilisateurs de drogues". On the left, the individual steps in the Bio-YODIE pipeline are visible. Bio-YODIE is available on GitHub.\footnote{https://github.com/GateNLP/Bio-YODIE}

\subsection{Coverage}

One approach to improving performance is accommodating more synonyms. This is challenging, however, as a more permissive approach may make more mistakes, and some ways of accommodating variation may be slow to execute. In Bio-YODIE a new approach was trialled. The UMLS, being a thesaurus, contains many alternative ways of expressing the same concept, offering untapped potential for synonym acquisition. We trialled an experimental approach. A script was prepared that took alternative labels for the same entity in UMLS, pairwise, and matched up the words contained in them. Where words were similar but not identical (as determined using a threshold on length-adjusted Levenshtein distance) they were considered synonyms. Having paired and removed these matching words, the remainder was also considered synonymous. Consider for example "pancreatic cancer" and "neoplasm, pancreas". "Pancreatic" and "pancreas" become synonyms, leaving "cancer" to match "neoplasm", which also gets added as a synonym. In some cases, a redundant word becomes synonymous with the empty string; these are also added. Scores are assigned to synonym pairs based on the frequency with which that synonym appeared, adjusted for the number of alternative labels for that CUI; CUIs with many labels tend to inflate the count disproportionately.

Thresholds were applied to the automatically derived synonyms. These were tuned on the MIMIC 2014 training data. A threshold of 12 was chosen for most synonyms; however where an item is to be replaced with the empty string, a high threshold of 1000 was imposed. An example of such an item is the ending "NOS", meaning "not otherwise specified", which can usually be dropped from a term without a change in its meaning. However, generally speaking, removing words from terms is somewhat perilous, resulting as it does in something more ambiguous, so therefore a high threshold was chosen, reducing the list of items that may be discarded to just eleven items, most of which are the bracketed semantic types that often appear at the end of a term, as in for example "headache (finding)". A manual edit of the synonyms was then performed to remove erroneous and problematic items; for example "declined" and "decline" can have quite different connotations. The final synonym list had 7097 items. Having applied this to the gazetteer, an additional 54644 items were included, constituting around 2\% of the total; permutations were applied only to the shorter terms, that are more likely to be matched in text, so this small addition has a disproportionate impact on the result without substantially increasing RAM requirements at runtime.  Figure~\ref{table:synexamp} gives twelve items taken from the top of the synonym list, as a representative illustration. Its coverage of abbreviations, plural forms, capitalization variation, adjectival forms and spelling mistakes is typical of the work.

\begin{table}
\begin{center}
\resizebox{.4\textwidth}{!}{%
\begin{tabular}{|r|l|}
\hline
\textbf{From} & \textbf{To} \\
\hline
inflammatory & inflammat \\
(presence) serum or plasma & serpl ql \\
antigen concentration point time unspecified & (presence) ag \\
Indomethacin & indomethacin \\
doudenal & duodenal \\
steatoses & steatosis \\
sutura & suture \\
musculus & muscle \\
arteriosis & arteriosus \\
sphincters & sphincter \\
audiometric & audiometry \\
darbepoetin & darbopoetin \\
\hline
\end{tabular}
}
\end{center}
\caption{Representative examples of automatically derived synonyms}
\label{table:synexamp}
\end{table}

Increasing recall without impacting negatively on precision has been hard to achieve so far in this domain. However, manual inspection of the output reveals that often the spuriously annotated items are quite sensible, though not matching exactly what was given in the corpus. Where spurious and incorrect items are semantically close to the gold standard answer, prioritizing recall makes sense. In order to investigate coverage in the light of this observation, a small section of the MIMIC 2014 corpus was prepared with additional annotations for adjectival forms, and where a mention in text could be linked validly to several CUIs, all were included. Results show that the data-driven approach of deriving synonyms from the UMLS that we trial here has benefits over linguistic variation in expanding semantically coherent coverage. Although we were not able to demonstrate an improvement on the evaluation corpus, we believe the approach has merit, and future work will involve a more thorough evaluation before including it in Bio-YODIE.

\subsection{Disambiguation}

Main approaches to named entity linking involve applying candidates to text on the basis of string matches. The main areas therefore by which disambiguation performance might be improved are through better matching of strings and better selection of the right candidate from among several. It is this latter area on which we focus in this section. Information that might be used to select the best match can be divided into two types; those that don't evaluate candidates on the basis of the context in which the mention appears, and those that do. The former category might be considered prior probabilities of one form or another. The latter we refer to as context clues.

\subsubsection{Priors}

A prior probability for the likelihood of a CUI occurring is a valuable information source. Essentially, we wish to know how important this entity is, how widely mentioned. Such a prior probability can be obtained in a number of ways. For example, in the general domain, the number of incoming links to the Wikipedia page for an entity gives a measure of its popularity and results in an extremely strong baseline. A second prior probability of interest is the likelihood of a particular term referring to a particular entity. For example, does "OD" more commonly indicate "overdose" or "osteochondritis dissecans"? This will be referred to in this work as the "link probability". In the biomedical domain, manually linked corpora are scarce, but possibilities do exist to obtain these priors. The two we have found the most successful are considered here. Prior probabilities are also convenient in that they can be calculated in advance, requiring no run-time processing, which would slow down the system.

\paragraph{Corpus Prior}

Annotated corpora can be easily used to calculate frequencies for CUI mentions. Furthermore, where a fully manually linked corpus is available, link probabilities can be obtained. The MIMIC corpus offers a training set of 299 documents, which can be used to calculate priors. The SLAM data has been split into two parts, one of which, in conjunction with the MIMIC training portion, has been used to create a set of prior probabilities, which are then used as scores in the system. Corpus priors appropriate to the use case have the potential to improve performance considerably--more than any other method tested--but must be created by the user with their own data. Bio-YODIE offers the possibility to drop in these priors.

\paragraph{Co-occurrence Graph}

Previous work has demonstrated the value of graph disambiguation approaches such as PageRank in making use of information such as the MRREL table in UMLS, which provides information about how entities relate to each other (for example they might be narrower or broader concepts etc.). Personalized PageRank has been turned to the task of making a final joint disambiguation; however, it improves only a little on PageRank~\cite{agirre2010graph}, and PageRank can be calculated over the graph in advance, making it faster at runtime. In calculating a full PageRank for the whole graph, we in essence create a new kind of prior; how connected is this entity? How important is it in the context of other entities? The fact that Personalized PageRank improves only a little on this perhaps emphasizes how difficult a good prior probability is to beat. Including a PageRank score provides a way to make use of any available graph information.

We calculated a (static) PageRank over the MRCOC (co-occurrences) table in UMLS. This table gives concept co-occurrences in PubMed articles, which are gold-standard. We used only those since 2000, and only those that occur more than once.

\subsubsection{Context Clues}

Prior probabilities give an indication of how likely a concept is to appear, but they don't make use of information in the surrounding document about what concept might appear in this context. Various approaches to evaluating a candidate in context have been trialled in the context of YODIE and other work. Bio-YODIE provided an opportunity to try something new. The recent surge of research into neural approaches to word representation has demonstrated the potential in replacing words with vectorial embeddings to bring in value from large volumes of unlabeled data and exploit co-occurrence to create richer representations of words. These representations might then be compiled to create vectors for entities and context, that can be compared for congruence. We made use of word2vec~\cite{le2014distributed} embeddings calculated over PubMed, PubMed Central and Wikipedia biomedical topics.\footnote{From here: http://bio.nlplab.org/} We calculated a vectorial representation for the entities by adding word vectors from their abstracts. We calculated a vector representation for the context of the mention, again, by adding word vectors. These vectors were then compared using cosine to produce a score giving an indication of the goodness of fit of that entity in context. Evaluation revealed however that the approach contributed little. It was also notable that the system became much slower when the context scoring was in place, since calculations are necessarily performed at runtime. This experience echoed that of previous work in the general domain~\cite{gorrell2015using}, in demonstrating that prior probability is an excellent source of information that it is hard to add anything to, and that work using context clues, whilst appealing, incur a high cost with regards to speed. It also raises questions however about the suitability of the data to demonstrate an improvement in this area.

\section{Evaluation}

MetaMap and MetaMapLite have been chosen for comparison systems for the evaluation because they are publicly available, popular systems. Another strong contender would be the cTAKES system; however the approach taken to named entity linking in this system is highly similar to that taken by MetaMapLite, offering a similar range of coverage extension functionality in addition to basic dictionary lookup, and no disambiguation, making it redundant for our purposes (cTAKES also contributes a wide range of further language processing functionality but that is not under consideration here). Both MetaMap and MetaMapLite were wrapped in GATE plugins in order to perform the experiments conducted for this work. The plugins are available under open source licenses~\footnote{https://gate.ac.uk/userguide/sec:misc-creole:metamap}~\footnote{https://github.com/GateNLP/gateplugin-MetaMapLite}. Further recent evaluation work, including cTAKES, has been shared by Demner-Fushman \textit{et al}~\shortcite{demner2017metamap}.

\subsection{MetaMap}

The basis of MetaMap's entity linking is a complex named entity recognition approach. Having found entities, near matches are then sought in the UMLS. The linguistically grounded approach to finding and permuting the mentions in text seems likely to add intelligence, but the cost is high in terms of runtime processing. MetaMap uses SICStus Prolog to provide a microservice architecture to parse text, identify mentions and locate matches in the UMLS. Extensive linguistically motivated permutation of the string in text is performed in order to broaden the matches to entity labels in the UMLS and achieve a higher recall. Linguistic cues are also used to assign a score to the quality of the match, which is then used for disambiguation. Parameters can be used to select the semantic types of interest and the UMLS distribution to use among other things.

\subsection{MetaMapLite}

Recently, the MetaMapLite system has emerged from the same team~\cite{demner2017metamap}, offering faster processing times, and transitioning to the "push" approach. The gazetteer approach is linguistically void, though some permutations have been precalculated, and the resulting system lacks the linguistically based disambiguation approach offered in MetaMap. Disambiguation is unaddressed at the time of writing.~\footnote{https://metamap.nlm.nih.gov/MetaMapLite.shtml} It is written in Java and provides an API. It includes a gazetteer, which it uses to find the matches in text. A configuration file makes it possible to specify a range of parameter preferences.

\subsection{Corpus}

The MIMIC II data\footnote{https://mimic.physionet.org/} offers three corpora manually annotated with CUIs for a subset of the UMLS semantic types. Two test corpora contain 133 and 100 documents respectively, and a training corpus of 299 documents is also available. This corpus is available on request to researchers meeting the application requirements. The patient notes include discharge summaries and ECG and echo reports, providing a rich and varied sample of medical text in American English. It has been used here to perform the evaluation.

\subsection{Results}

Table~\ref{table:results} shows that in terms of processing time, the two ``push'' approaches, MetaMapLite and Bio-YODIE, outperform MetaMap, as expected given their much simpler processing at runtime. The question is, can this speed improvement be achieved without loss of accuracy and coverage? The answer it seems is yes. In terms of accuracy, both Bio-YODIE and MetaMapLite outperform MetaMap by a couple of percent, with Scott's Pi confirming the result. In terms of recall (coverage) and precision, however, we see differing merits to the systems. Bio-YODIE achieves the better recall by several percent. However, MetaMapLite has the substantially higher precision of all three systems. In terms of F1, Bio-YODIE and MetaMapLite have the superior result, with MetaMapLite ahead by just a fraction of a percent. In summary, simpler approaches have shown their merit in this case.

\begin{table}
\begin{center}
\resizebox{\columnwidth}{!}{%
\begin{tabular}{|l|l|l|l|l|l|l|}
\hline
 & Secs & Prec L & Rec L & F1 L & Acc & Scott's Pi\\
\hline
MetaMap & 3811 & 0.574 & 0.568 & 0.571 & 0.857 & 0.856 \\
MetaMapLite & 986 & \textbf{0.654} & 0.549 & \textbf{0.597} & 0.877 & 0.876 \\
Bio-YODIE & \textbf{573} & 0.582 & \textbf{0.605} & 0.593 & \textbf{0.883} & \textbf{0.882} \\
\hline
\end{tabular}
}
\end{center}
\caption{MIMIC results for the three systems on 2014 test data}
\label{table:results}
\end{table}

\section{Discussion}

As we have seen above, in the general domain, "push" systems show stability and robustness, but don't naturally identify unknown entities, requiring additional machinery to add this functionality if it is required. Historically, "pull" systems have received more attention because named entity recognition is the more established field, presenting an initially more manageable task that is useful in the context of automated database population. Many named entity linking systems arose from this existing work. However, this assumption is limiting in this case.

The most notable difference between the three systems reviewed, as shown in the previous section, is with respect to coverage; the systems differ in their balance between precision and recall. Since it is hard to beat prior probabilities with regards to improving disambiguation, coverage improvements offer the most potential for improving overall performance. There is plenty of scope for improving recall.

Future work should involve evaluation on additional corpora, since it is not clear the extent to which idiosynchracies in this corpus might have influenced the result presented here. We also suggest that gold standard corpora for named entity linking should include all CUIs that might match a mention rather than just picking one. We note that small corpora make it hard to discern the impact of improvements. We draw the reader's attention to earlier work~\cite{tissot2015analysis} showing the difficulty experienced by annotators in consistently following complex annotation manuals; in the medical field there can be many edge cases.

\bibliographystyle{aaai}
\bibliography{bio-yodie.bib}




\end{document}